\documentclass[sigconf, nonacm]{acmart}

\usepackage{hyperref}
\usepackage{url}
\usepackage{booktabs}       
\usepackage{amsfonts}       
\usepackage{nicefrac}       
\usepackage{microtype}      
\usepackage{xcolor}         
\usepackage{graphicx}
\usepackage{subfigure}
\usepackage{amsmath}
\usepackage{algorithm}
\usepackage{algorithmic}
\usepackage{multirow}
\usepackage{diagbox}
\usepackage{enumitem}

\newtheorem{theorem}{Theorem}[section]

\AtBeginDocument{%
  \providecommand\BibTeX{{%
    \normalfont B\kern-0.5em{\scshape i\kern-0.25em b}\kern-0.8em\TeX}}}

\begin{document}

\title[Farthest Greedy Path Sampling for Two-shot Recommender Search]{Farthest Greedy Path Sampling for \\Two-shot Recommender Search}

\author{Yufan Cao}
\email{caoyf20@mails.tsinghua.edu.cn}
\affiliation{%
  \institution{Tsinghua University}
  \streetaddress{30 Shuangqing Rd}
  \city{Haidian Qu}
  \state{Beijing Shi}
  \country{China}}

\author{Tunhou Zhang}
\email{tunhou.zhang@duke.edu}
\affiliation{%
  \institution{Duke University}
  \streetaddress{Durham, NC 27708}
  \city{Durham}
  \state{North Carolina}
  \country{USA}}

\author{Wei Wen}
\email{wewen@meta.com}
\affiliation{
    \institution{Meta AI}
    \city{Menlo Park}
    \state{California}
    \country{USA}
}

\author{Feng Yan}
\email{fyan5@uh.com}
\affiliation{
    \institution{University of Houston}
    \city{Houston}
    \state{Texas}
    \country{USA}
}

\author{Hai Li}
\email{hai.li@duke.edu}
\affiliation{
  \institution{Duke University}
  \streetaddress{Durham, NC 27708}
  \city{Durham}
  \state{North Carolina}
  \country{USA}}

\author{Yiran Chen}
\email{yiran.chen@duke.edu}
\affiliation{
  \institution{Duke University}
  \streetaddress{Durham, NC 27708}
  \city{Durham}
  \state{North Carolina}
  \country{USA}}

\renewcommand{\shortauthors}{Cao, et al.}

\begin{abstract}

Weight-sharing Neural Architecture Search (WS-NAS) provides an efficient mechanism for developing end-to-end deep recommender models. However, in complex search spaces, distinguishing between superior and inferior architectures (or \textit{paths}) is challenging. This challenge is compounded by the limited coverage of the supernet and the co-adaptation of subnet weights, which restricts the exploration and exploitation capabilities inherent to weight-sharing mechanisms. To address these challenges, we introduce Farthest Greedy Path Sampling (FGPS), a new path sampling strategy that balances path quality and diversity. FGPS enhances path diversity to facilitate more comprehensive supernet exploration, while emphasizing path quality to ensure the effective identification and utilization of promising architectures. By incorporating FGPS into a Two-shot NAS (TS-NAS) framework, we derive high-performance architectures. Evaluations on three Click-Through Rate (CTR) prediction benchmarks demonstrate that our approach consistently achieves superior results, outperforming both manually designed and most NAS-based models.
\end{abstract}



\maketitle

\section{Introduction}

\begin{figure*}[h]
    \centering
    \includegraphics[width=\linewidth]{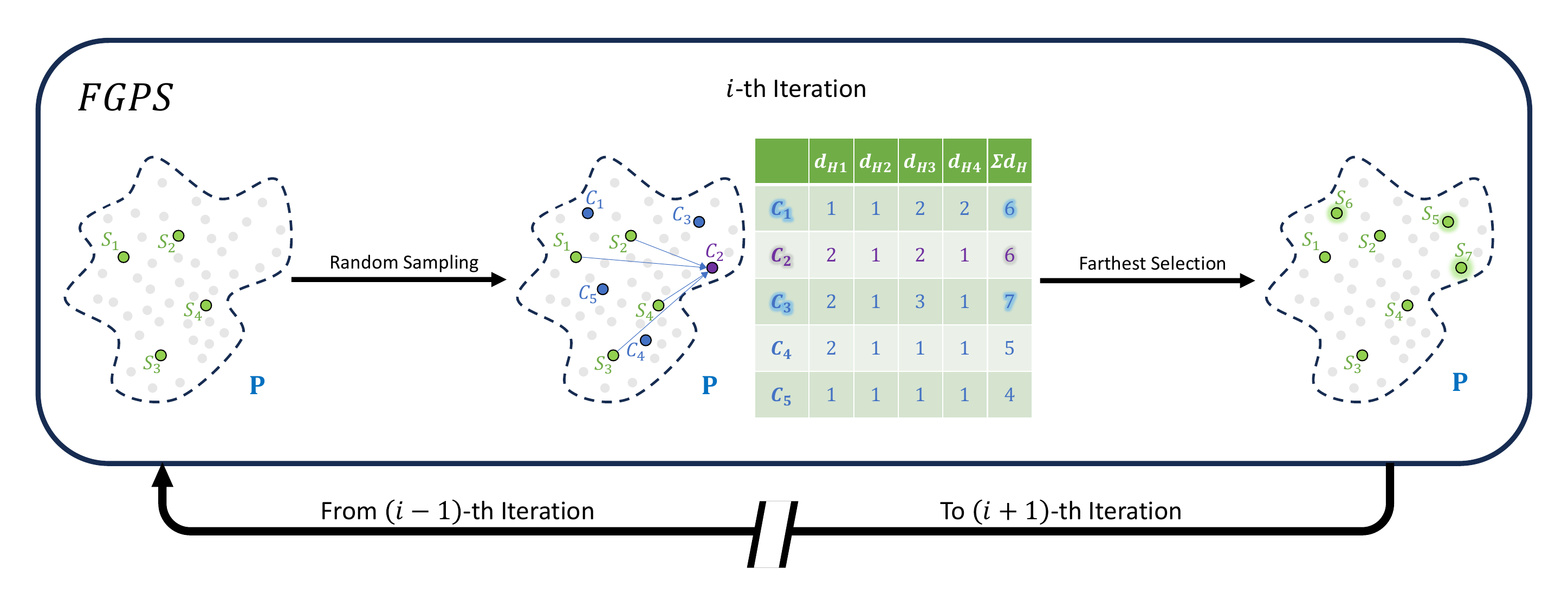}
    \caption{Illustration of Farthest Greedy Path Sampling (FGPS). With the selected path set (green dots) after the $(i-1)$-th iteration, FGPS samples paths (blue dots) from a path set $\mathbf{P}$ (gray dots) and pick out the farthest ones with largest total distance from the selected paths and add them to the selected set as the result of the $i$-th iteration.}
    \label{fig:fgps}
\end{figure*}

Recommender systems, essential in domains such as entertainment~\cite{bell2007lessons}, social media~\cite{koren2009matrix}, and e-commerce~\cite{ricci2010introduction}, have continually evolved to refine user experiences by predicting preferences more accurately. 
Historical lines of approaches to improve precision include the utilization of collaborative filtering~\cite{sarwar2001item, koren2009matrix}, content-based filtering~\cite{mooney2000content, pazzani2007content}, and knowledge-based systems~\cite{burke2002hybrid, tso2008tag}. These methodologies have been pivotal in tailoring recommendations by analyzing and modeling distinct user features, thus facilitating more precise estimations of user behavior.
In the evolution of recommender systems, the integration of Deep Neural Networks (DNNs) results in a notable paradigm shift. Initial DNN strategies~\cite{guo2017deepfm, naumov2019deep, lian2018xdeepfm, song2019autoint} heavily rely on manual, hand-crafted efforts. The emergence of Neural Architecture Search (NAS)~\cite{song2020towards, dnas, zhang2022nasrec} enables a more systematic and efficient exploration of the DNN design space, reducing the reliance on manual intervention.

Weight-sharing NAS (WS-NAS)~\cite{yu2020bignas,cai2019once} achieves remarkable architectural flexibility among various NAS methodologies.
WS-NAS substantially reduces computational costs by employing a single \textit{supernet} to represent the entire search space. 
A supernet is a large, all-encompassing model, from which we zero out certain components or connections to sample a \textit{subnet} that corresponds to a \textit{path} from input to output. The process of sampling a subnet, i.e., a path from the supernet is therefore termed ``path sampling''. Weights of the supernet can be naively copied to expedite the evaluation of sampled subnets, based on which NAS identifies presumably top-performing architectures. WS-NAS successfully achieves remarkable architectural flexibility while substantially reducing computational costs.

This methodology opens up opportunities for NAS with highly complex search space settings and prohibitive computational costs. However, it suffers from the ``co-adaptation'' problem arising from the gap between weights of the supernet and the optimal weights of a subnet yielding its real performance~\cite{optgap}. With such an optimization gap, the incorrectly evaluated top-performing architectures may be unintentionally eliminated during the search, resulting in worse performance of top-performing candidates. We attribute part of this problem to relatively naive existing path sampling strategies~\cite{Chu_2021_ICCV, guo2020single} where few consider coverage of top-performing paths.

In this paper, we propose ``Two-shot Deep Recommender Search'' (TS-NAS), a novel NAS approach that advances the path sampling strategy to improve the ranking correlation of different subnets. 
TS-NAS performs a search on the WS-NAS supernet twice to obtain the optimal subnets.
In the first shot, we follow the practice of NASRec~\cite{zhang2022nasrec} to train a supernet and identify a set of well-performing paths, from which we sample a path subset with a maximized diversity measured by Shannon Entropy. 
The sampled top-performing paths have the ``farthest'' pairwise distances and ``greediest'' performance, hopefully having good coverage of the supernet for their diversity, thereby mitigating operator imbalance and weight co-adaptation during search. 
In the second shot, we train a supernet on Farthest Greedy Paths, after which we perform an evolutionary search for top-performing architectures.
We name this path sampling strategy ``Farthest Greedy Path Sampling'' (FGPS, Figure~\ref{fig:fgps}) and highlight it as the major contribution of this work.

We build our approach on the broad but challenging NASRec search space with heterogeneous building operators, dense connectivity, and elastic dimensions. 
We thoroughly examine our method on three Click-Through Rates (CTR) prediction benchmarks Criteo Kaggle, Avazu, and KDD Cup 2012. Results demonstrate that our method outperforms all hand-crafted models, and shows an improvement of $\sim$~0.001 in AUC compared to NAS-crafted models in Criteo and KDD benchmarks. We summarize our contribution as follows:  
\begin{itemize}[noitemsep,leftmargin=*]

    \item We design a heuristic path sampling approach named ``FGPS'', which yields a set of paths boasting both diversity 
    and coverage of top-performing architectures in the search space. This set of paths empirically enables the second shot of NAS to find better top-performing architectures.
    \item We propose a novel NAS approach for recommender systems named ``TS-NAS''. It enables training the supernet with paths sampled by FGPS and more accurate evaluation of top-performing recommender system architectures. 
    \item The results show that the architectures our method finds can achieve state-of-the-art (SOTA) performance in most main CTR prediction tasks, verifying its ability to automatically explore and exploit an intricate design space with high heterogeneity.
\end{itemize}

\section{Related Works}

\noindent \textbf{Deep Learning-based Recommender Systems.} 
Deep Neural Networks (DNNs) serve as the backbone models that support end-to-end inference in recommender systems, such as Click-Through Rate (CTR) prediction applications~\citep{covington2016deep, he2017neural, sedhain2015autorec}.
Starting with Wide \& Deep learning~\citep{cheng2016wide}, the integration of dense and sparse features has played a crucial role in modeling user preferences and optimizing click-through rates. Techniques like Factorization Machines~\citep{guo2017deepfm}, Attention~\citep{song2019autoint}, crossing layers~\citep{wang2021dcn,wang2017deep}, and instance-guided masks~\citep{wang2021masknet} have been developed to enhance explicit and implicit feature interactions, thereby establishing strong design priors to drive DNN designs. Another line of research focuses on optimizing the set of feature interactions in recommender systems, including attention/field-aware factorization machines~\citep{xiao2017attentional, juan2016field}, feature interaction search~\citep{liu2020autofis, gao2021progressive}.
More recent research~\citep{song2020towards,zhang2022nasrec} utilize Neural Architecture Search (NAS) to expand the scope of exploration in deep recommenders, such as dense connectivity, heterogeneity, and multimodality. Yet, studies on the quality of NAS, such as ranking quality, are rarely conducted in existing works.

\noindent \textbf{Path Sampling in One-shot Search.} 
One-shot NAS~\citep{cai2019once,yu2020bignas} utilizes a supernet to represent the search space and carry path sampling to select different set of subnets during training. In one-shot NAS, the quality of path sampling decides the ability of a NAS algorithm to distinguish between superior and inferior architectures~\cite {bender2018understanding}. Single-path sampling~\citep{guo2020single}, sandwich rule~\citep{yu2020bignas}, progressive shrinking~\citep{cai2019once}, and greedy path sampling~\citep{you2020greedynas,huang2022greedynasv2} are past approaches on computer vision problems. Yet, existing approaches do not consider supernet coverage during path selection. This is because the search space within computer vision is relatively smaller and simpler, containing only a single type of block (e.g., MobileNet-v2).

\section{Methodology}

We base our search space on the settings of NASRec~\cite {zhang2022nasrec}, which is a composition of building operators, connectivity, and elastic dimension.

\begin{figure*}[h]
    \centering
    \includegraphics[width=\linewidth]{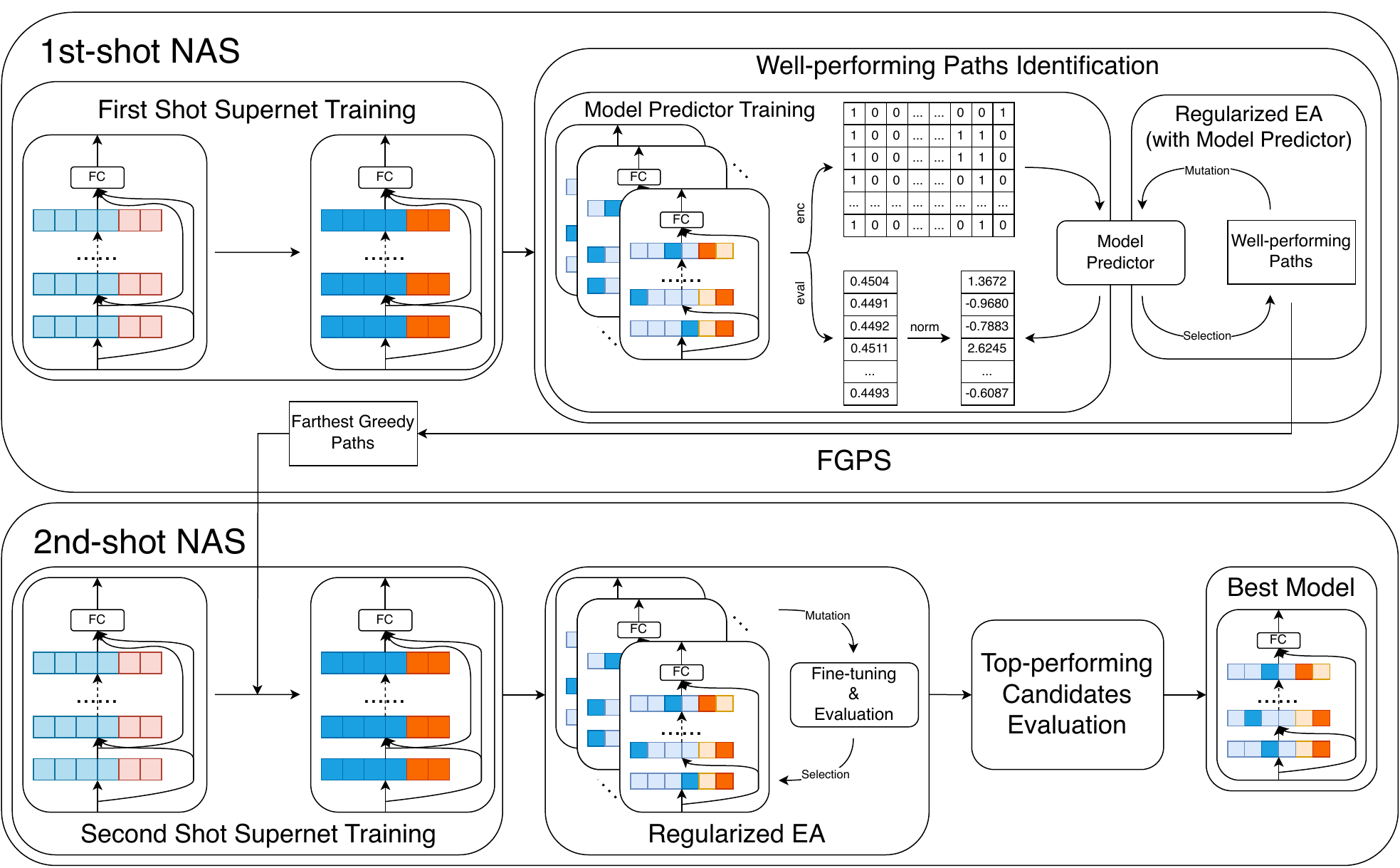}
    \caption{Overview of Two-Shot Neural Architecture Search with Farthest Greedy Path Sampling(FGPS). FGPS prepares paths for training the supernet in the second shot, mitigating the co-adaptation problem and leading to a better best model.}
    \label{fig:workflow}
\end{figure*}

\subsection{Supernet Configuration and its Sampled Path Encoding}\label{encoding}

Following the definition in~\cite{zhang2022nasrec}, a supernet $\mathcal{S}$ with choice blocks from $1$ to $N$ is a tuple of connections $\mathcal{C}=[C^{(i)}]_{i=1}^N$, operators $\mathcal{O}=[O^{(i)}]_{i=1}^N$, and dimensions $\mathcal{D}=[D^{(i)}]_{i=1}^N$ over $N$ choice blocks. $C^{(i)}$ defines the connectivity between choice block $j$ and choice block $i$ for all $j < i$ ($0$ as input), avoiding loops. For example, if there is a connection between choice block $1$ and choice block $3$, the outputs of choice block $1$ should be taken as part of the inputs of choice block $3$. $O^{(i)}$ defines the set of building operators in choice block $i$. And $D^{(i)}$ defines the dimension settings in choice block $i$. 

A valid path (i.e., a subnet) $\mathcal{P}$ = ($\mathcal{C}_{p}$, $\mathcal{O}_{p}$, $\mathcal{D}_{p}$) from input to output within the supernet is a tuple of sampled sets of connections, operators, and dimensions from the search space for each choice block. We formulate the encoding of a sampled path as follows: 

\begin{itemize}[noitemsep,leftmargin=*]
    \item On the inter-block level, we use zero-one encoding to represent the connectivity among blocks. In block $j$, operator $o$ within a search space of $N$ blocks, the encoding $\epsilon(\mathcal{P}, j, o)\in \{0,1\}^N$, and $\epsilon[i] = \mathbf{1}_{<i,j>\in C_{p}^{(j)}}$. For example, the fourth choice block out of seven that takes in raw input, the outputs of the first and the third block as its inputs correspond to an encoding of ``1101000''. The connectivity is encoded sequentially for all modules and then concatenated to form the inter-block-level encoding.
    \item On the intra-block level, we pre-assign options for operators, and structural parameters (e.g.\#dimensions, with or without dense/sparse interaction), and encode in-block choices with their indices. Encoding is sequentially done for all blocks, concatenated and appended to inter-block-level encoding. We discuss intra-block details in Section \ref{4_1}.
\end{itemize}

\subsection{Measurement of Distance and Entropy among Paths}
We formulate \textit{distance}(how different one path is from another), and \textit{entropy}(how diverse a path set is) for navigation towards better exploration and exploitation within our encoding space.
The path encoding above constructs a bijection between the architecture of a subnet in the search space, and a vectorized representation in the 280-dimensional encoding space. We measure \textit{distance} between paths $\mathcal{P}_m$ and $\mathcal{P}_n$ with Hamming distance~\cite{hammingdist} $\texttt{Dist}_H$:

\begin{equation}
    \texttt{Dist}_H(\mathcal{P}_m, \mathcal{P}_n) = \#\texttt{different\_symbols}(\texttt{enc}(\mathcal{P}_m), \texttt{enc}(\mathcal{P}_n)).
\end{equation}

$\texttt{Dist}_H$ evaluates how \textit{far} path $\mathcal{P}_m$ is from $\mathcal{P}_n$, allowing the selection of more diverse paths during sampling. Hamming distance is employed with the assumption that different operators hold equivalent chances of achieving high final performances. 

Similarly, to measure the diversity of a set of paths $\mathbf{P} = \{\mathcal{P}_m\}_{m=1}^{k}$, we exploit Shannon entropy~\cite{shannonent}. The Shannon entropy $\texttt{Ent}_S$ of $\mathbf{P}$ is formulated as follows:

\begin{equation}
    \texttt{Ent}_S(\mathbf{P}) = \sum_{\texttt{pos}=1}^{\texttt{len}(\texttt{enc})}\left[\sum_{e=1}^{\#\texttt{enc}(\mathbf{P})[\texttt{pos}]}-\texttt{f}_\texttt{pos}(e)\log\texttt{f}_\texttt{pos}(e)\right]  ,
\end{equation}
where $\#\texttt{enc}(\mathbf{P})[\texttt{pos}]$ denotes the number of different symbols on the $\texttt{pos}$-th dimension, and\begin{equation}
    \texttt{f}_{\texttt{pos}}(e) = \sum_{m=1}^k \mathbf{1}(\mathcal{P}_m\left[\texttt{pos}\right]=e) / k
\end{equation}
is the frequency of symbol $e$ on the $\texttt{pos}$-th dimension of the encoding. We propose:

\begin{theorem}\label{th3_1}
    Given a path set $\mathbf{P} = \{\mathcal{P}_m\}_{m=1}^{k}$ and an encoding function $\texttt{enc}$, a maximum Shannon entropy on a path subset $\mathbf{P}^* = \{\mathcal{P}^*_n\}_{n=1}^{r} \subseteq \mathbf{P}$ can be approximated if
    \begin{equation}
        \mathbf{P}^* = \arg\max_{\mathbf{P}'} \left[\sum_{i=1}^{r} \sum_{j=i+1}^{r} \texttt{Dist}_H(P'_{i}, P'_{j})\right] .
    \end{equation}
\end{theorem}

The theorem above indicates that maximizing the pairwise Hamming distance of sampled paths improves the Shannon entropy of the sampled subset, making them more diverse.

\subsection{Farthest Greedy Path Sampling}

\begin{algorithm}
\caption{Farthest Greedy Path Sampling}\label{algo_fgps}
\begin{algorithmic}
\REQUIRE A path set $\mathbf{P}$, an encoding method $\texttt{enc}$, \#budget $B$ per iteration, \#paths $K$ added per iteration.
\STATE Selected path set $\mathbf{P}_S \gets B$ different random paths sampled \\from $\mathbf{P}$
\STATE Candidate path set $\mathbf{P}_C \gets \varnothing$
\WHILE{One iteration takes reasonable time}
    \WHILE{$\lvert\mathbf{P}_C\rvert < B$}
        \STATE Randomly sample a path $\mathcal{P}$ from $\mathbf{P}$
        \IF{$\mathcal{P} \notin \mathbf{P}_S\cup\mathbf{P}_C$}
        \STATE Add $\mathcal{P}$ to $\mathbf{P}_C$
        \ENDIF
    \ENDWHILE
    \FOR{$\mathcal{P}_i$ in $\mathbf{P}_C$}
        \STATE Compute the sum of Hamming distance between $\mathcal{P}_i$ and all the paths in $\mathbf{P}_S$
    \ENDFOR
    \STATE Pick out $K$ paths from $\mathbf{P}_C$ with highest distance summation and add them to $\mathbf{P}_S$
    \STATE $\mathbf{P}_C \gets \varnothing$
\ENDWHILE
\RETURN $\mathbf{P}_S$

\end{algorithmic}
\end{algorithm}

Previous works of weight-sharing one-shot NAS~\cite{yu2020bignas,cai2019once} suffer from the so-called ``co-adaptation'' problem, which arises directly from the design of the method~\cite{optgap}. In each step of training the supernet, they first sample a path randomly from the supernet and perform a one-step gradient descent on the related parameters in the supernet. In the downstream architecture search session, the parameters of related parts of the supernet is directly copied to give a rough evaluation of the performance of a subnet. Though this method efficiently cuts down computational burdens, in evaluation there is an inevitable gap between the supernet parameters and the optimal parameters which yields the real performance of a subnet. 

It is impossible to address the co-adaptation problem as the once-trained parameters in the supernet can never be optimal for all sampled paths, so we aim to reduce its impact only on top-performing subnets, with the insight that NAS pays little attention to poor-performing architectures. 


We introduce ``farthest greedy path sampling'' (FGPS) as a foundation to practice the above insight. We confine the path set $\mathbf{P}$ in Theorem \ref{th3_1} as a well-performing path set, and aim to find a path subset $\mathbf{P}^*$ with high diversity quantified by Shannon entropy. In FGPS, we use pairwise path distance as a proxy to maximize the Shannon entropy and iteratively sample new paths that are most different from the ones already sampled. The detailed sampling strategy is described in Algorithm~\ref{algo_fgps}. As the bridge between first-shot and second-shot NAS, FGPS explores the well-performing path set to gain diversity, and exploits sampled paths for a better model search. Paths sampled by FGPS are referred to as Farthest Greedy Paths.


\subsection{Weight-sharing Two-shot NAS Workflow}

Equipped with FGPS, we propose a novel approach of NAS, ``Weight-sharing Two-shot NAS''(TS-NAS). The TS-NAS workflow can be summarized as below:

\begin{enumerate}[leftmargin=*]
    \item Train first-shot supernet with randomly sampled paths;
    \item Use an efficient way (see Section \ref{modelpred}) to roughly identify a set of well-performing paths;
    \item Perform FGPS on the identified path set to find a diverse well-performing path subset to acquire Farthest Greedy Paths;
    \item Train second-shot supernet with Farthest Greedy Paths;
    \item Perform evolutionary search on sampled paths from trained second-shot supernet to generate top-performing path candidates;
    \item Evaluate top-performing candidates and pick out the best architecture.
\end{enumerate}

With supernet training on a well-performing diverse path set, parameter updating is directed towards the optimal parameter of top-performing paths, narrowing the aforementioned parameter gap. Therefore, it mitigates the ``co-adaptation'' problem of the top-performing paths.

\subsection{Identification of a Well-performing Path Set}\label{modelpred}

With a search space containing $\sim10^{33}$ different architectures, it is impossible to identify a large set of well-performing paths with direct evaluation of the subnets. To address this challenge, we adopt the encoding space. 

Given a 1-shot trained supernet, we randomly sample a lot of paths $\mathbf{P}_\mathrm{rand}$ from it, fine-tune them (which is computationally affordable), and get an evaluation $\texttt{logloss}(\mathbf{P}_\mathrm{rand})$. Then we perform $\texttt{enc}$ to map these sampled paths into the encoding space, generating $\texttt{enc}\mapsto\texttt{logloss}$ mappings inside $\mathbf{P}_\mathrm{rand}$. We then train a simple MLP with embedding to learn the mappings. As the encoding function has constructed a mapping from paths to encodings, such method constructs a direct mapping from model architectures to their predicted performance and effectively lowers the computational budgets.

With this mapping from paths to predicted performances as a proxy, we can now identify a large set of well-performing paths with ease. We use an evolutionary algorithm to iteratively generate new paths and keep the top ones with the highest predicted performance.


\section{Experiments}

We first delineate our basic experimental settings including search space and dataset configuration, hyperparameters and baseline implementations. Then we provide a detailed step-by-step Two-shot NAS Workflow in very fine grain, followed by a comparative analysis of experimental results with baselines. Finally we investigate how FGPS improves Shannon Entropy during path sampling, and the role of Shannon Entropy throughout the workflow. 

\subsection{Experimental Setup}\label{4_1}

\subsubsection{Search Space and Datasets} Our method is built upon the open-source implementation provided by NASRec\cite{zhang2022nasrec}, from which we adopt the defined search space.

At a macroscopic level, the search space consists of seven choice blocks, permitting arbitrary connections from preceding blocks to subsequent ones. At a microscopic level, the functionality within each block is governed by specific operators. These operators can be categorized based on their input and output types:

\begin{itemize}[noitemsep,leftmargin=*]
    \item Operators handling dense inputs to produce dense outputs: This category includes the Fully-Connected (FC) layer, Sigmoid Gating (SG) layer, and Sum layer.
    \item Operators processing sparse inputs to yield sparse outputs: This category includes the Embedded Fully-Connected (EFC) layer and the Attention (Att) layer.
    \item Operators managing a combination of dense and sparse inputs to generate dense outputs: This category includes the Dot-Product (DP) layer.
\end{itemize}

In scenarios where multiple operators are concurrently active within a dense or sparse branch, their outputs are aggregated via summation. In our implementation, the DP layer has been modified with an expansion of the dense dimension by a factor of 16.

Within each choice block, mergers exist to integrate "dense" and "sparse" data. Specifically, a "dense-to-sparse" merger accepts dense outputs, processes them via a Fully Connected (FC) layer, and then reshapes the outcome into a 3D sparse tensor. This tensor is subsequently concatenated to the sparse outputs. Conversely, a "sparse-to-dense" merger takes as input sparse outputs, processes them using a Factorization Machine (FM)~\cite{rendle2010factorization}, and then adds the resulting output to the dense outputs.

We perform a Neural Architecture Search and model evaluation on three click-through-rate (CTR) prediction benchmarks: Criteo Kaggle\footnote{https://www.kaggle.com/c/criteo-display-ad-challenge}, Avazu\footnote{https://www.kaggle.com/c/avazu-ctr-prediction}, and KDD Cup 2012\footnote{https://www.kaggle.com/c/kddcup2012-track2}. We used a split of 80\% training data, 10\% validation data, and 10\% data for each benchmark. We use Stratified K-fold~\cite{scikit-learn} to obtain the dataset splits with a random seed of 2018, ensuring the consistency with existing works such as AutoInt~\cite{song2019autoint}, AutoCTR~\cite{song2020towards}, and NASRec~\cite{zhang2022nasrec}.

All model performances are assessed using Logarithmic Loss as the evaluation metric.

\subsubsection{Training Settings.} To ensure a fair comparison and better demonstrate our efforts in architecture search, we employ a consistent training protocol for all architectures, deliberately avoiding hyperparameter tuning. Specifically, we employed the Adagrad optimizer~\cite{duchi2011adaptive} with an initial learning rate of 0.12. A batch size of 256 was utilized for the Avazu and Criteo datasets, while a size of 512 was designated for the KDD dataset. All models were optimized over one single epoch, and a cosine learning rate schedule~\cite{loshchilov2016sgdr} is added to gradually decay the learning rate to 0 at the end of the training epoch.

\subsubsection{Baselines} We implement all baselines based on open-source code and paper descriptions. Most models have 2 layers in dense processing layers and 6 layers in overarching layers, as recommended in AutoCTR~\cite{song2020towards}.
We carefully tune the architecture of each baseline to ensure a competitive implementation over all CTR benchmarks. 

\begin{table*}[h]
    \begin{center}
    \caption{Performance of TS-NAS with FGPS on Three Click-Through Rate Prediction tasks.}
    \vspace{0.5em}
    \scalebox{1.0}{
    \begin{tabular}{|c|c|cc|cc|cc|}
    \hline
     & \multirow{2}{*}{\textbf{Method}} & \multicolumn{2}{c|}{\textbf{Criteo}}  &  \multicolumn{2}{c|}{\textbf{Avazu}} & \multicolumn{2}{c|}{\textbf{KDD}}  \\
      & & logloss & AUC & logloss & AUC & logloss & AUC \\
    \hline \hline
    \multirow{7}{*}{\textbf{Manually-designed Models}} 
    & DLRM & 0.4408 & 0.8105 & 0.3760 & 0.7864 & 0.1528 & 0.7980  \\
    & DeepFM~ & 0.4402 & 0.8111 & 0.3798 & 0.7797 & 0.1541 & 0.7905 \\
    & xDeepFM & 0.4400 & 0.8114 & 0.3777 & 0.7836 & 0.1544 & 0.7886  \\
    & AutoInt+ & 0.4447 & 0.8062 & 0.3796 & 0.7803 & 0.1499 & 0.8120  \\
    & SerMaskNet & 0.4408 & 0.8104 & 0.3782 & 0.7828 & 0.1496 & 0.8128  \\
    & ParaMaskNet & 0.4402 & 0.8111 & 0.3772 & 0.7843 & 0.1496 & 0.8130  \\
    & DCN-v2 & \underline{0.4389} & \underline{0.8125} & 0.3789 & 0.7814 & 0.1548 & 0.7864   \\
    \hline

    \multirow{3}{*}{\textbf{NAS-crafted Models}}
    & NASRec@Small & 0.4391 & 0.8123 & \underline{0.3751} & \underline{0.7880} & 0.1489 & 0.8165 \\
    & NASRec@Full & 0.4394 & 0.8119 & \textbf{0.3749} & \textbf{0.7883} & \underline{0.1487} & \underline{0.8168} \\
    & TS-NAS w/FGPS@Full & \textbf{0.4387} & \textbf{0.8127} & 0.3755 & 0.7874 & \textbf{0.1485} & \textbf{0.8179} \\
    
    \hline
    \end{tabular}
    \label{tab:ctr_results}
    }
    \end{center}
    \vspace{-1em}

\end{table*}

\begin{itemize}[noitemsep,leftmargin=*]
    \item \textbf{DLRM}~\cite{naumov2019deep} has a two-tower architecture composed of 2 MLP layers in dense feature processing and a 6-layer MLP in high-order feature processing. All hidden layers have 1024 units.

    \item \textbf{DeepFM}~\cite{guo2017deepfm} has a wide \& deep architecture that integrates the benefits of FM for factorization and deep learning for abstract feature learning. The dense feature processing is done with a two-layer MLP, and a 6-layer MLP is for high-order feature extraction. All hidden layers have 1024 units.
    
    \item \textbf{xDeepFM}~\cite{lian2018xdeepfm} is a two-tower model featuring a Compressed Interaction Network (CIN) to perform vector-wise explicit feature generation. Besides CIN, xDeepFM has 2 MLPs in dense feature processing and 6 MLPs in high-order feature processing. All hidden layers have 1024 units.
    
    \item \textbf{AutoInt+}~\cite{song2019autoint} extends the AutoInt architecture by incorporating additional techniques such as feature selection and self-attention mechanisms to enhance model performance, particularly in handling high-dimensional feature spaces. All hidden layers have 1024 units.
    
    \item \textbf{SerMaskNet}~\cite{wang2021masknet} employs a series of masking techniques within a deep neural network structure and across MaskBlocks to enable more efficient and effective feature interaction and selection, fostering better learning and generalization. SerMaskNet has three MaskBlock in series with all hidden layers having 1024 units.
    
    \item \textbf{ParaMaskNet}~\cite{wang2021masknet} operates similarly to SerMaskNet but introduces parallel masking networks to capture different kinds of features for a nuanced pattern capturing. ParaMaskNet has three MaskBlock in parallel with all hidden layers having 1024 units.
    
    \item \textbf{DCN-v2}~\cite{wang2021dcn} stands for Deep and Cross Network version 2, an evolution of the original DCN, featuring enhanced deep and cross combination for more powerful feature interaction learning and representation. All hidden layers have 1024 units.
    
    \item \textbf{NASRecNet}~\cite{zhang2022nasrec} leverages weight-sharing NAS within the realm of recommender systems and employs ``Single-operator Any-connection'' sampling, evolutionary algorithms to approximate the best model. We follow the open-source implementation for this baseline.
\end{itemize}

\subsection{Two-Shot NAS Workflow}

We use the Criteo dataset as a representative example and explain our workflow through each phase.

\subsubsection{First-shot Supernet Training.} During the initial phase of the first-shot NAS, the supernet undergoes training using paths sampled at random. For the first 15,000 steps, there is a progressively diminishing probability that the entire supernet is trained to avoid model collapse. We adopt the ``Single-operator Any-connection'' strategy~\cite{zhang2022nasrec} for this sampling approach. In detail, each choice block, we select one dense operator and one sparse operator, and we permit connections between all blocks. This strategy is specifically tailored for the subsequent model predictor training phase for its superior performance that the ranking of model predictions closely aligns with their actual value rankings. For the training process, we employ the Adam optimizer~\cite{kingma2014adam} with an initial learning rate of 0.001 and a batch size of 512.


\subsubsection{Model Predictor Training} After training the supernet, we pick paths from it at random to create a set denoted as $\mathbf{P}_R$. To identify a large set of well-performing paths efficiently, we train a model predictor using this set.

First, we apply the encoding function $\texttt{enc}$ on the path set and acquire $\texttt{enc}(\mathbf{P}_R)$ as a matrix containing the tokenized path configurations. Then, we prepare the models for these paths using the weights from the trained supernet and fine-tune them by training for another 0.5K steps. This gives us a dataset that shows how different encoded paths perform in terms of Log Losses.

For each path, we embed each token in the encoding into 15-dimensional vectors with a look-up table. After combining these vectors, we process them using a Multi-Layer Perceptron~\cite{rumelhart1986learning}. The goal is to adjust the model predictor so that its predictions match the normalized Log Losses of the paths. We choose the best hyperparameter settings for the model based on the test Kendall's $\tau$ coefficient~\cite{kendall1938new}. Finally, we train the model using the entire dataset to prepare it for the next steps.

\subsubsection{Identification of a Well-performing Path Set} Utilizing the model predictor as a surrogate, we efficiently identify a vast collection of high-performing paths using a regularized Evolutionary Algorithm (EA)~\cite{holland1992adaptation, goldberg2013genetic}. We begin by initializing a population of 20K paths. In each iteration, 51 paths are randomly sampled from this population. The top-performing path from this sample is then selected and mutations are applied to generate 64 child paths. Based on their predicted normalized log losses, the top 10 child paths are chosen. This process is repeated for 20K generations, leading to a final set of 200K paths. Without the model predictor, obtaining such a large set of well-performing paths would be prohibitively expensive.

\subsubsection{Farthest Greedy Path Sampling} From the well-performing path set, we begin by eliminating redundant encodings. Due to the long-tailed distribution of the predicted performance, we also exclude the bottom $\sim$ 7.5\% of the paths. Starting with an initial set of 200 paths selected, we draw 200 paths from those not yet visited during each iteration. From this drawn subset, we retain the 20 paths that have the greatest total distance from all paths in our selected set. This procedure results in $\sim$ 5.5K distinct paths after iterations, namely Farthest Greedy Paths. The role of Shannon Entropy in this selection process is further explored in section \ref{sec_fgps}. At this point, we are poised to commence the second shot of our Neural Architecture Search.

\subsubsection{Second-shot Supernet Training} In the second-shot training, we utilize the Farthest Greedy Paths to train the supernet. Throughout the initial 15,000 steps, we maintain a warm-up process for the supernet, with the warm-up probability gradually reducing to zero. To ensure a fair comparison with baseline methods and to emphasize the significance of FGPS, we use the AdaGrad optimizer~\cite{duchi2011adaptive} with an initial learning rate of 0.12, consistent with the settings of the baselines.

\subsubsection{Second-shot Evolutionary Algorithm} In the second shot of the regularized evolutionary algorithm, we aim for a more detailed assessment of the subnets. Each subnet undergoes fine-tuning for 0.5K steps and is then evaluated on the test dataset. The process begins by initializing a population comprising 128 subnets. From a sample of 64, the best-performing model is identified, and 8 mutations from this model are subsequently evaluated. This procedure is carried out over 240 generations, after which the outcomes are sorted based on their evaluated losses in the test dataset.

\subsubsection{Top-performing Candidates and Best Model Evaluation} We focus on the top 15 subnets identified by their minimal test losses, and further fine-tune and evaluate them. The model exhibiting the lowest test loss is then selected as the best. Subsequently, we assess this best model starting from scratch, employing a batch size of 256, the AdaGrad optimizer, and an initial learning rate of 0.12 for one epoch. This approach is consistent with the evaluation configurations of the baselines. The resulting losses are presented in Table~\ref{tab:ctr_results}.

\subsection {Comparative Analysis}




Our approach significantly surpasses the majority of manually designed recommender systems across all datasets. It also consistently outperforms models discovered through NAS, with a stable margin of log loss observed on the Criteo and KDD datasets. Additionally, our method achieves an improvement of approximately 0.001 in the AUC metric when compared to the baselines. This underscores our proposition that FGPS mitigates the optimization gap and co-adaptation problem, offering superior performance to one-shot weight-sharing NAS for recommender systems without the need for parameter tuning. This can also be corroborated in Section \ref{5_1}.

An interesting observation is our method's inability to exceed NASRec's performance on the Avazu dataset. We postulate that this discrepancy arises from the Avazu dataset's unique characteristic of lacking dense inputs. This results in limited interactions between dense and sparse components, which are precisely where the search space's heterogeneity is most demonstrated. Consequently, our search algorithm might become trapped in a local optimum, incorporating redundant dense/sparse interactions. 
We have scrutinized the top paths identified by the second-shot EA to corroborate this conjecture in Section~\ref{5_3}.

\subsection {FGPS and Shannon Entropy}\label{sec_fgps}

\begin{figure}[b]
    \centering
    \includegraphics[width=\linewidth]{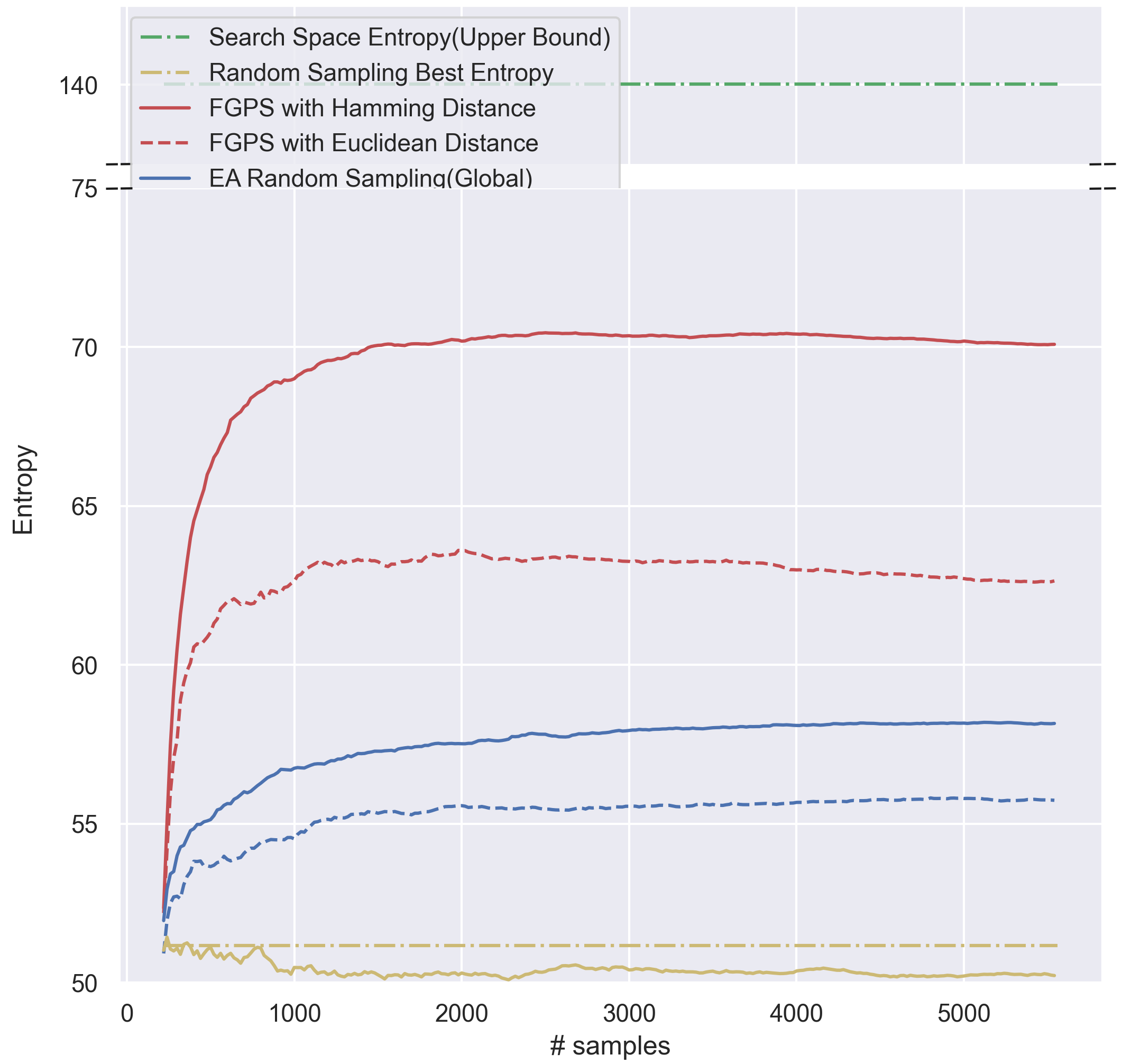}
    \caption{Comparison of Shannon Entropy of path sets iteratively sampled according to different strategies.}
    \label{ablation-ent}
\end{figure}

We take the NAS on Criteo dataset in the full NASRec~\cite{zhang2022nasrec} search space as an example and investigate Farthest Greedy Path Sampling.

\subsubsection{Shannon Entropy Improves with FGPS}

To validate the efficacy of FGPS in enhancing path set diversity measured by set Shannon Entropy, we benchmarked it against several path sampling baseline strategies. In the EA random sampling, paths are sampled randomly multiple times, retaining those with the highest local/global entropy of the sample/selected set. Figure~\ref{ablation-ent} portrays a clear distinction in performance between FGPS and its counterparts, with FGPS manifesting superior sampling efficacy. Notably, FGPS utilizing Hamming Distance surpasses its Euclidean Distance counterpart, echoing that a uniform categorical distribution is a more apt surrogate optimum for Shannon Entropy. All heuristic sampling methods significantly outperform random sampling, validating their statistical significance.

Using a sampling strategy as FGPS with Hamming Distance, the Shannon Entropy of the selected path set rapidly increases to approximately 70, and terminates with $\sim$~5.5K paths sampled, attesting the effectiveness of FGPS. As the sampling progresses, the entropy of the FGPS-sampled path set slightly decreases, suggesting an efficient attainment of optimal Shannon Entropy by FGPS at an early sampling stage.

\subsubsection{Shannon Entropy throughout the Workflow} 

\begin{figure}[b]
    \centering
    \includegraphics[width=\linewidth]{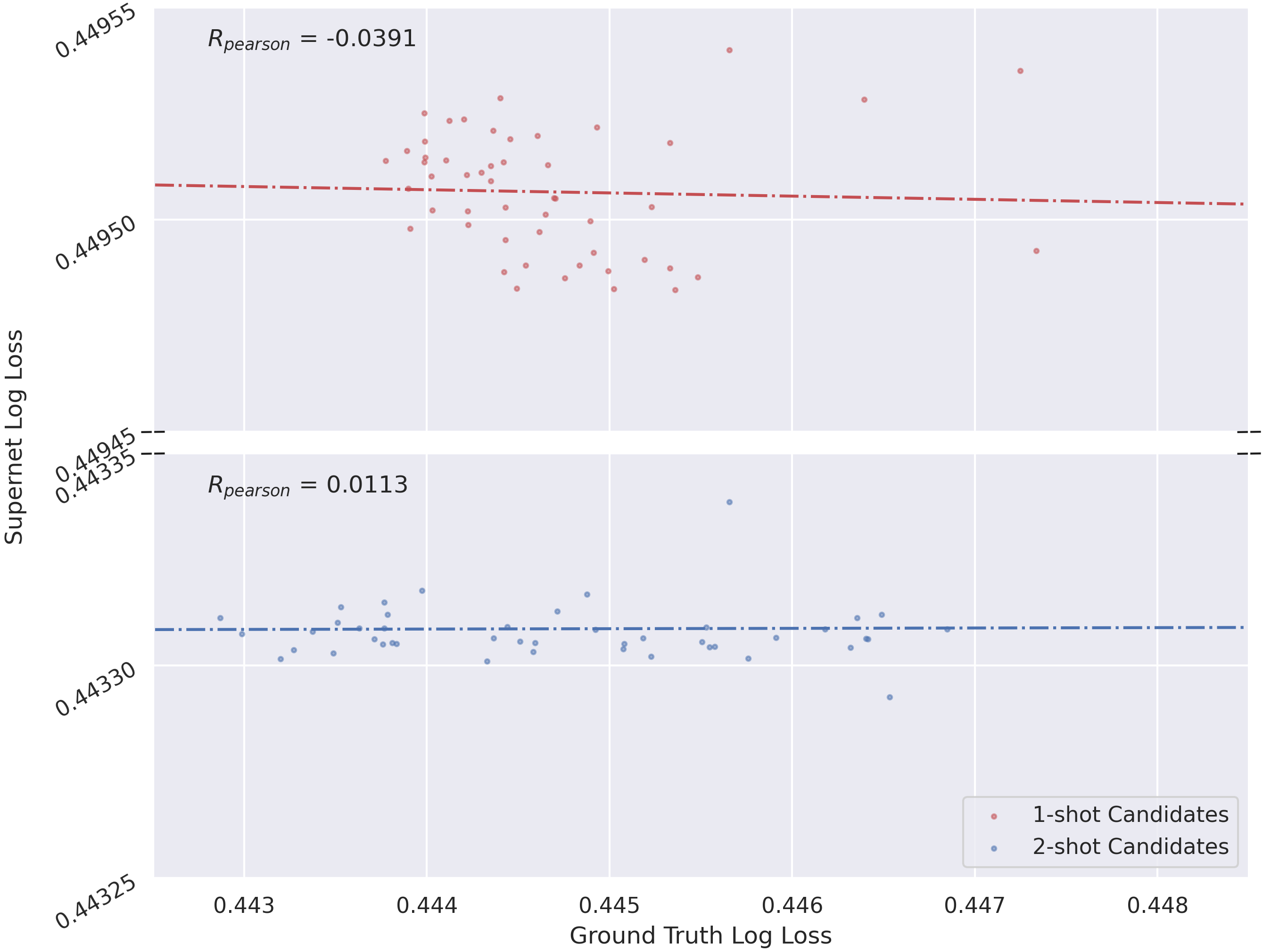}
    \caption{Ranking analysis of top-performing paths found by 1-shot Weight-sharing NAS and TS-NAS w/FGPS.}
    \label{ablation-ranking}
\end{figure}
Using the encoding approach described in Section \ref{encoding}, we calculate the maximum Shannon Entropy for our search space to be 140.0. This value corresponds to a set of paths that has a uniform distribution for each position in the encoding vector. After applying the first-shot Regularized Evolutionary Algorithm, the resulting path set has an entropy of approximately 50.5. We attribute this decrease in Shannon Entropy to the search space narrowing down to a biased set of paths that perform better. Given this, FGPS's goal is to elevate the entropy, ensuring a diverse set of paths even within a performance-biased path set. This allows for a broader exploration of the supernet, while focusing on the top-performing subnets for a better exploitation of them in the second shot NAS.

\section{Discussion}

We provide the ranking analysis, visualization of top-performing paths for distribution analysis and review the top-performing paths.

\subsection{Ranking Analysis}\label{5_1}

We investigate the top 50 subnets found by the evolutionary algorithm by 1-shot weight-sharing NAS and TS-NAS with FGPS to verify that FGPS improves the rankings among the top-performing architectures. We denote Supernet Loss as the test loss after fine-tuning the parameters in the trained supernet, and Ground-Truth Loss as the test loss after training the subnet from scratch. We observe an evident gap between the supernet losses, and an improvement in Pearson correlation in Figure \ref{ablation-ranking}. Considering the high sensitivity of the top-performing architectures, this attests that our method mitigates the optimization gap for the top-performing models. Moreover, the ground-truth performance distribution of top candidates discovered by TS-NAS with FGPS exhibits an evident shift towards lower losses compared to one-shot results, proving the effectiveness of FGPS for TS-NAS.

\subsection{Top-performing Paths Visualization}
\begin{figure}[b]
    \centering
    \includegraphics[width=\linewidth]{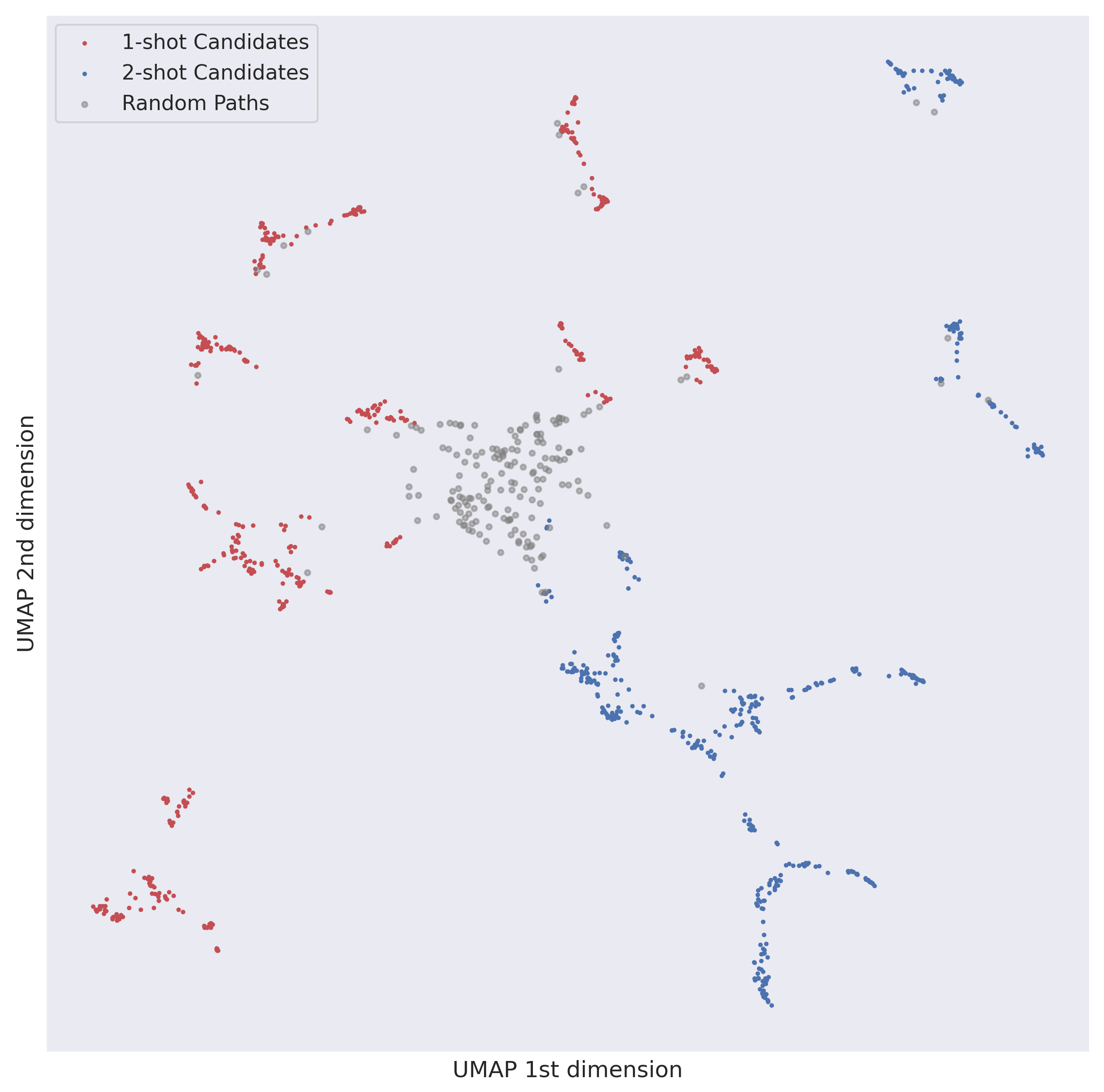}
    \caption{UMAP plot of path encodings of first-shot, second-shot candidates and randomly sampled 200 paths. A lower overlap with random paths suggests a higher statistical significance.}
    \label{ablation-pathvis}
\end{figure}

We compare the top candidates found by the regularized evolutionary algorithm in the first shot and the second shots. We randomly sample ~200 paths, and fit a UMAP~\cite{mcinnes2020umap} together with first-shot and second-shot candidates in Figure~\ref{ablation-pathvis}. We observe a much higher overlap between random paths and 1-shot candidates, suggesting a higher statistical significance of paths found by TS-NAS. We also observe that candidate paths found by TS-NAS lie on a more continuous manifold, while candidate paths found by 1-shot NAS fall into many discrete clusters. This confirms the ability of TS-NAS with FGPS to exploit what has been explored in the first shot and focus on the most promising paths in the search space.

\subsection{Top-performing Model Architecture}\label{5_3}



Compared with the best model found by 1-shot weight-sharing NAS in Criteo dataset on the same search space, our method finds a model with much more effective modules. Specifically, only one out of the 14 operators is unused in our model, in stark contrast to the six unused operators observed in the best model from the 1-shot weight-sharing NAS. 


Our method did not achieve superior performance in the relatively simpler Avazu dataset, which lacks dense inputs and necessitates minimal communication between dense and sparse features. Upon examining the top models we discovered, we observed that the majority employed Sigmoid Gating for over half of the dense operators. This led to an abundance of redundant or ineffective dense feature processing. This limitation can be attributed to the model predictor's challenge in semantically discerning each operator and a Farthest Greedy Path set that overly prioritizes dense/sparse interactions. That our method roughly made par with NASRec~\cite{zhang2022nasrec} in Avazu suggests that employing FGPS might not be essential in a less heterogeneous search space. Instead, the strength of FGPS comes when exploring more diverse and intricate search spaces.



\subsection{FGPS for TS-NAS as a Generic WS-NAS Paradigm}


The inherent challenge of coadaptation is deeply rooted in the design of Weight-sharing NAS and is not exclusive to Recommender Systems. We propose FGPS for TS-NAS as a universal framework tailored for weight-sharing NAS, especially when dealing with a search space characterized by high heterogeneity and complexity.

The true power of FGPS lies in its capacity to balance the dual objectives of exploration and exploitation. This dual capability makes FGPS a pivotal component in navigating the intricate balance between exploration and exploitation in NAS search space.

\section{Conclusion}



In this study, we propose a new path sampling strategy ``Farthest Greedy Path Sampling''(FGPS), and a two-shot Neural Architecture Search(TS-NAS) workflow tailored for deep recommender systems. TS-NAS incorporates FGPS to adeptly balance path quality and diversity, thus maximizing the effectiveness of the search process. Through rigorous evaluations on multiple Click-Through Rate (CTR) prediction benchmarks, our approach consistently showcased superior performance against all manually designed and most NAS-derived recommender models. This work not only advances the state-of-the-art in NAS for recommenders, introduces a new paradigm of weight-sharing NAS in highly heterogeneous search spaces, but also sets a new benchmark for future research in this domain.

\bibliographystyle{ACM-Reference-Format}
\bibliography{main}

\end{document}